\newif\ifanonymousversion
\anonymousversionfalse
\ifanonymousversion
  \PassOptionsToClass{anonymous}{acmart}
\fi
\documentclass[sigconf,nonacm,10pt]{acmart}

\usepackage{booktabs}
\usepackage{array}
\usepackage{multirow}
\usepackage{xcolor}
\usepackage{algorithm}
\usepackage{algpseudocode}
\usepackage[normalem]{ulem}
\usepackage{enumitem}
\usepackage{placeins}

\AtBeginDocument{%
  }

\setcopyright{none}
\acmDOI{}
\acmISBN{}
\newcommand{\maybeincludegraphics}[2]{%
\IfFileExists{#1}{\includegraphics[width=#2]{#1}}{%
\fbox{\parbox[c][0.24\textheight][c]{#2}{\centering Upload figure file:\\ \texttt{\detokenize{#1}}}}}}

\newcommand{\pmstd}[1]{{\scriptsize\,$\pm$\,#1}}

\newif\ifshowchanges
\showchangesfalse
\definecolor{revisionred}{RGB}{190,0,0}
\DeclareRobustCommand{\rev}[1]{\ifshowchanges\textcolor{revisionred}{#1}\else#1\fi}
\newenvironment{revision}{\ifshowchanges\color{revisionred}\fi}{}
\AtBeginDocument{\pdfstringdefDisableCommands{\def\rev#1{#1}}}

\begin{document}

\title[How Much Graph Is Too Much?]{How Much Graph Is Too Much? Graph Construction and Exposure in Spatio-Temporal Forecasting}

\ifanonymousversion
  \author{Anonymous Authors}
  \renewcommand{\shortauthors}{Anonymous Authors}
\else
  \author{Saj Maru}
  \affiliation{\country{}}
  \email{sajmaru360@gmail.com}
  \author{Qiyu Han}
  \affiliation{\country{}}
  \email{han541@purdue.edu}
  \author{Krishna Mamidala}
  \affiliation{\country{}}
  \email{krishna.reddy985@gmail.com}
  \renewcommand{\shortauthors}{Maru et al.}
\fi

\begin{abstract}
\begin{revision}
Spatio-temporal graph neural networks combine a node's history with information from graph neighbors. The central question is whether graph neighbors provide predictive information beyond each node's own history. Forecast accuracy depends on both the choice of neighbors and the way graph messages enter the temporal model. We study these choices as graph construction and graph exposure on U.S. state unemployment, traffic speed, and electricity demand. Our diagnostic design uses two comparisons: varying exposure while holding the domain graph fixed, and varying construction while holding late gated integration fixed. A GRU using only each node's own history and randomized graph controls help assess what the graph contributes. Under a common change forecasting protocol, the same domain graph can improve or worsen error relative to the GRU across integration designs. Late integration alone is insufficient, while adding a fusion gate yields lower mean error in all three tasks. With late gated integration fixed, a graph based on residual correlations has the lowest mean error among the tested graphs, although randomized graphs perform similarly on unemployment. The findings support evaluating graph construction and integration together, with conclusions tied to the comparisons actually performed.
\end{revision}

\end{abstract}

\maketitle
\pagestyle{plain}
\thispagestyle{plain}

\section{Introduction}
\label{sec:intro}

\begin{figure}[t]
    \centering
    \includegraphics[width=\columnwidth]{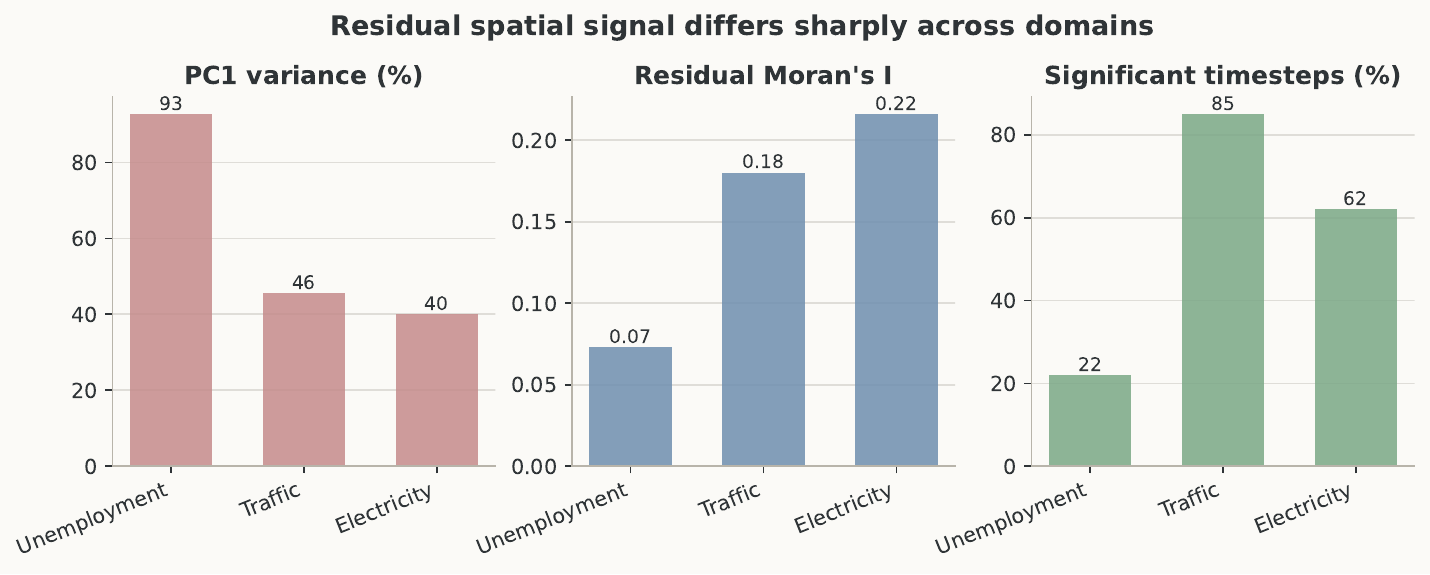}
    \caption{\rev{\textbf{Shared variation and residual spatial dependence.} PC1 is the variance share of the first principal component. Residual Moran's $I$ measures spatial autocorrelation after common component removal \cite{moran1950notes}. The right panel reports the share of timesteps for which it is significant.}}
    \label{fig:intro_diagnostic}
\end{figure}

\begin{revision}
Spatio-temporal graph neural networks (STGNNs) forecast related time series by passing information between nodes on a graph \cite{li2018dcrnn,yu2018stgcn,bai2020agcrn}. Road proximity, state borders, and electricity grid ties can supply plausible graphs for sensors or regions. We study when these graphs improve forecasts beyond a model that uses each node's own history. This requires assessing what information the neighbors add and how the temporal model uses it. A graph can describe a meaningful relation without providing useful additional information for the forecast target. Many graph forecasting methods start with a given graph and assume that its neighbor information helps. Our problem is to test this assumption by separating the choice of neighbors from the way the model uses their information.

We distinguish two design choices. \emph{Graph construction} selects the neighbors that exchange information. \emph{Graph exposure} determines where their messages enter the temporal model, how often graph operations are applied, and whether a gate controls their contribution. A poor forecast can have different explanations: the selected neighbors may provide little useful information, or the integration design may make poor use of their messages. An accuracy comparison that changes both choices at once cannot distinguish these explanations. This motivates comparisons that hold one choice fixed while varying the other.

Observed spatial dependence adds another difficulty. Similar movements across nodes may reflect common factors, seasonality, or shared shocks, some of which are already visible in a node's own history. Figure~\ref{fig:intro_diagnostic} illustrates the contrast across our datasets. A dominant common component explains 93\% of unemployment variation, while traffic and electricity retain more spatial dependence after common component removal. These descriptive statistics motivate studying different dependence patterns. They do not establish whether the graph improves a forecast, which we assess against a baseline using only each node's own history and against randomized graphs.
\end{revision}

\begin{revision}
We investigate construction and exposure on U.S. state unemployment, traffic speed, and electricity demand. All comparisons use a common protocol for predicting standardized changes and evaluating reconstructed levels, defined in Section~\ref{sec:forecasting-task}. The study makes three contributions:
\begin{itemize}[leftmargin=*,topsep=3pt,itemsep=3pt]
    \item \textbf{A diagnostic design.} We define construction and exposure as separate choices. One comparison varies exposure with each domain graph fixed; the other varies construction with late gated integration fixed. This supports conclusions conditional on the fixed choice.
    \item \textbf{Evidence that integration changes graph value.} The same domain graph can yield errors above or below the GRU baseline. On electricity, late integration without a fusion gate has 9.21\% WMAPE, compared with 5.27\% with the gate and 6.18\% for the GRU. Delaying integration alone is therefore insufficient in this setting.
    \item \textbf{Evidence that topology needs explicit controls.} With late gated integration fixed, the residual correlation graph has the lowest mean error among the tested graphs. Randomized graphs perform similarly on unemployment, indicating that gains from adding graph messages need not establish the value of the intended topology.
\end{itemize}
\end{revision}

\FloatBarrier
\section{\rev{Problem Formulation and Diagnostic Design}}
\label{sec:problem-form}

\begin{revision}
We separate two choices in graph forecasting. \emph{Graph construction} selects the neighbors available to the model. \emph{Graph exposure} describes where graph messages enter the model, how often they are applied, and whether a separate gate controls their contribution. We define both choices below and evaluate them through two comparisons with a shared forecasting protocol.
\end{revision}

\subsection{\rev{Forecasting Task and Training Protocol}}
\label{sec:forecasting-task}

\rev{We study forecasting over nodes $V=\{1,\ldots,N\}$. Node $i$ has value $y_{t,i}$ at time $t$. Given a window $X_{t-L+1:t}$ of standardized observations and an optional adjacency matrix $A\in\mathbb{R}^{N\times N}$, the model predicts the change at horizon $H$. Algorithm~\ref{alg:change-target} gives the common training target and the reconstruction of forecasts in the original units. It defines the task shared by every model. It ensures that comparisons of graph construction and graph exposure use the same target and the same reconstruction procedure.}

Section~\ref{sec:intro} shows that raw cross-node dependence can be dominated by common temporal factors. As a result, standard level forecasting is less diagnostic of graph value. If you predict raw future values, a graph model can appear deceptively useful by spreading information that all nodes already share, rather than learning genuine node-to-node predictive structure. We therefore train all neural models with the standardized change target in Algorithm~\ref{alg:change-target}.

\begin{algorithm}[H]
\caption{Standardized change forecasting target}
\label{alg:change-target}
\small
\begin{algorithmic}[1]
\Require Data $Y=\{y_{t,i}\}\in\mathbb{R}^{T\times N}$, graph $A$, lookback $L$, horizon $H$
\State Estimate node-wise training statistics $\mu_i$ and $\sigma_i$ for each node $i$
\State Standardize each series: $z_{t,i}\leftarrow (y_{t,i}-\mu_i)/\sigma_i$
\For{each forecasting window ending at time $t$}
    \State Form input $X_{t-L+1:t}=\{z_{\tau,i}:$\newline
    $\tau=t-L+1,\ldots,t;\ i=1,\ldots,N\}$
    \For{each node $i=1,\ldots,N$}
        \State Define target change $\Delta z_{t+H,i}\leftarrow z_{t+H,i}-z_{t,i}$
    \EndFor
    \State Predict all node changes $\Delta \hat z_{t+H}\leftarrow f_\theta(X_{t-L+1:t},A)$
    \State Train with $\mathcal{L}=\frac{1}{N}\sum_i \mathrm{SmoothL1}(\Delta \hat z_{t+H,i},\Delta z_{t+H,i})$
    \For{each node $i=1,\ldots,N$}
        \State Reconstruct $\hat z_{t+H,i}\leftarrow z_{t,i}+\Delta\hat z_{t+H,i}$
        \State Convert back $\hat y_{t+H,i}\leftarrow \mu_i+\sigma_i\hat z_{t+H,i}$
    \EndFor
\EndFor
\end{algorithmic}
\end{algorithm}

Algorithm~\ref{alg:change-target} makes the evaluation focus on whether the graph helps predict future changes, not whether it helps reconstruct shared levels. Under a level target, i.e., predicting the future value itself ($z_{t+H,i}$), neighboring nodes may carry redundant information about a dominant shared temporal component. Message passing can then reduce error by spreading this common state across nodes, even when the graph itself provides little useful predictive structure. The change target removes this easiest shortcut by anchoring each prediction at the node's own last observation.

To see why this matters, suppose the standardized series decomposes as $z_{t,i}=c_t+r_{t,i}$, where $c_t$ is a shared temporal component and $r_{t,i}$ is node-specific. A level target contains
\[
z_{t+H,i}=c_{t+H}+r_{t+H,i},
\]
so the loss can reward a graph model for predicting the shared component. The change target in Algorithm~\ref{alg:change-target} instead gives
\[
\Delta z_{t+H,i}
=
(c_{t+H}-c_t)+(r_{t+H,i}-r_{t,i}).
\]
Thus, the change target subtracts the common level at the forecast origin before the loss is computed. The graph is then tested on the harder and cleaner question of whether neighbor information helps predict future movement beyond the shared state already visible across nodes. This reframes message passing as a spatial inductive bias rather than a guaranteed source of signal. The rest of the study asks when this bias is predictive and when it instead injects misleading spatial aggregation into the temporal model.

\subsection{\rev{Graph Construction}}
\label{sec:construction}

\begin{revision}
Let $\mathcal{D}_{\mathrm{train}}$ denote the training observations and $A_{\mathrm{dom}}$ the domain graph. A construction rule $C$ produces the adjacency matrix
\begin{equation}
    A_C=C(\mathcal{D}_{\mathrm{train}},A_{\mathrm{dom}}).
\end{equation}
The five rules in Table~\ref{tab:graph-construction} compare domain structure, empirical correlation, and randomized topology. All correlations are estimated from the training period. The resulting graphs are fixed during model training and evaluation.

For the residual correlation graph, let $d_{t,i}$ denote the first difference used in graph estimation. A first difference is the change in a node's value from one time step to the next. Removing the mean difference leaves the part of that change that differs from the average change across nodes at the same time. We remove the mean difference across nodes at each time:
\begin{equation}
    r_{t,i}=d_{t,i}-\frac{1}{N}\sum_{j=1}^{N}d_{t,j}.
\end{equation}
We then compute correlations between the residual series and select the $k=5$ most positively correlated neighbors for each node. This removes the contemporaneous mean change before neighbor selection. It does not identify causal or directional predictive relationships.
\end{revision}

\begin{table}[t]
\ifshowchanges\color{revisionred}\fi
\centering
\caption{\rev{Graph construction rules and the question each comparison addresses.}}
\label{tab:graph-construction}
\small
\begin{tabular}{@{}>{\raggedright\arraybackslash}p{0.29\linewidth}>{\raggedright\arraybackslash}p{0.65\linewidth}@{}}
\toprule
\textbf{Graph} & \textbf{Construction and purpose} \\
\midrule
Domain prior
& State borders, road proximity, or grid ties. Tests the forecasting value of the domain graph. \\

Random degree matched
& Random neighbors with each node's outgoing degree preserved. Tests whether the identity of its neighbors matters. \\

Raw correlation kNN
& The $k=5$ neighbors with the highest positive correlations in the raw series. Tests neighbor selection based on shared levels. \\

Residual correlation kNN
& The $k=5$ neighbors selected after differencing and mean removal. Tests neighbor selection based on residual changes. \\

Rewired 100\%
& All domain edges rewired with graph density preserved. Tests the value of the original topology at the same density. \\

\bottomrule
\end{tabular}
\end{table}

\begin{table*}[t]
\ifshowchanges\color{revisionred}\fi
\centering
\caption{\rev{Graph exposure strategies. Frequency counts graph applications within one input window. The fusion gate combines local and graph representations.}}
\label{tab:coupling-methods}
\small
\begin{tabular}{@{}>{\raggedright\arraybackslash}p{0.18\linewidth}>{\raggedright\arraybackslash}p{0.16\linewidth}>{\raggedright\arraybackslash}p{0.10\linewidth}>{\raggedright\arraybackslash}p{0.10\linewidth}>{\raggedright\arraybackslash}p{0.35\linewidth}@{}}
\toprule
\textbf{Method} & \textbf{Insertion stage} & \textbf{Frequency} & \textbf{Fusion gate} & \textbf{Update sketch} \\
\midrule
GRU & None & None & No
& $H_t=\mathrm{GRUCell}(x_t,H_{t-1})$ \\
GCN$\rightarrow$GRU & Before temporal encoding & Each step & No
& $H_t=\mathrm{GRUCell}(M_A(x_t),H_{t-1})$ \\
T-GCN & Within GRU gates & Each step & No
& GRU gates use $M_A([x_t,H_{t-1}])$ \\
GRUCell$\rightarrow$GCN & After each memory update & Each step & No
& $H'_t=\mathrm{GRUCell}(x_t,H_{t-1})$,\newline $H_t=M_A(H'_t)$ \\
Late GRU$\rightarrow$GCN & After the full window & Once & No
& $H^T=\mathrm{GRU}(X_{t-L+1:t})$,\newline $H=M_A(H^T)$ \\
Late + gate & After the full window & Once & Yes
& Local and graph representations combined by Eq.~\ref{eq:late-gate} \\
\bottomrule
\end{tabular}
\end{table*}

\subsection{\rev{Graph Exposure}}
\label{sec:exposure}

\begin{revision}
Exposure is the integration strategy used for a given graph. We treat graph exposure as a methodological component. It specifies when the model computes graph messages and how it combines them with temporal representations. We distinguish the insertion stage, the frequency of graph operations within the input window, and the use of a separate gate to combine local and graph representations.

Write $M_A(U)=\mathrm{GCN}_A(U)$ for graph message passing on node features $U$. Table~\ref{tab:coupling-methods} defines five ways to combine it with a temporal encoder, alongside a GRU without graph messages. Early integration mixes the inputs before temporal encoding. Recurrent integration places graph operations inside the GRU gates or after each memory update. Late integration applies graph operations once, after encoding the full window.

For late gated fusion, the temporal encoder first produces $H^T$, with each node encoded from its own history. The graph representation $\widetilde H$ is then combined with $H^T$:
\begin{equation}
\begin{aligned}
    \widetilde H &= M_A(H^T),\\
    \alpha &= \mathrm{gate}([H^T,\widetilde H]),\\
    H &= (1-\alpha)\odot H^T+\alpha\odot\widetilde H.
\end{aligned}
\label{eq:late-gate}
\end{equation}
Here $\alpha$ denotes the learned fusion weights, $[\cdot,\cdot]$ denotes concatenation, and $\odot$ denotes elementwise multiplication. The separate fusion gate controls the local and graph contributions. It is distinct from the update and reset gates within a GRU. Comparing late integration with and without this gate tests whether a final graph operation alone is sufficient.
\end{revision}

\subsection{\rev{Diagnostic Comparisons}}
\label{sec:diagnostic-comparisons}

\begin{revision}
We use two comparisons. First, we fix the domain graph and compare all exposure strategies against the GRU baseline. This tests how the same graph performs under different integration designs. Second, we fix late gated fusion and vary graph construction. This tests which neighbors are useful under that integration design. Both comparisons use the same data splits, forecast targets, and evaluation metric.

These comparisons cover selected combinations of construction and exposure. They do not establish how every graph would perform under every integration strategy. The exposure variants also differ in their computations and parameters, so their differences cannot be attributed to timing alone. We report GAT, AGCRN, and CaST as additional reference models because they introduce their own graph mechanisms.
\end{revision}

\FloatBarrier
\section{Experiments}
\label{sec:experiments}

\rev{We apply the comparisons in Section~\ref{sec:diagnostic-comparisons} to unemployment, traffic, and electricity forecasting.}

\subsection{Setup}
\label{sec:setup}

Table~\ref{tab:datasets} summarizes the three datasets, which span distinct graph-signal regimes. Monthly unemployment rates are dominated by a common national factor, high-frequency traffic speeds follow a physical road network, and hourly electricity demand is shaped by regional grid and weather patterns. Each dataset represents a domain where spatial relationships are expected to matter---making all three natural candidates for graph-based forecasting. All data is split chronologically (70/15/15) to simulate real forecasting conditions.

\begin{table*}[ht]
\centering
\caption{Dataset summary. Three domains are selected to represent distinct graph-signal regimes.}
\label{tab:datasets}
\small
\begin{tabular}{@{}>{\raggedright\arraybackslash}p{0.18\linewidth}>{\raggedright\arraybackslash}p{0.245\linewidth}>{\raggedright\arraybackslash}p{0.245\linewidth}>{\raggedright\arraybackslash}p{0.245\linewidth}@{}}
\toprule
 & \textbf{Unemployment (BLS LAUS)} & \textbf{Traffic (METR-LA)} & \textbf{Electricity (EIA-BA)} \\
\midrule
What it measures  & Monthly unemployment rate per U.S. state & Vehicle speed on LA freeways & Hourly electricity demand per balancing authority \\
Why a graph seems natural & Interstate labor mobility links state labor markets & Congestion propagates along road segments & Power flows across AC grid interconnections \\
\midrule
Nodes             & 51 U.S. states               & 207 road sensors              & 52 balancing authorities \\
Domain graph      & 109 edges (state borders)  & 1,515 edges (road proximity)  & 204 edges (AC grid ties) \\
Frequency         & Monthly                    & 5-minute                      & Hourly \\
Time span         & Jan 2007--Dec 2024         & Mar--Jun 2012                 & Jan 2019--Apr 2026 \\
Timesteps         & 216                        & 34,272                        & 64,248 \\
Target variable   & Unemployment rate (\%)     & Traffic speed (mph)           & Electricity demand (MWh) \\
Lookback $\to$ horizon & 12 mo.\ $\to$ 1 mo. & 12 steps (1\,h) $\to$ 12 steps & 168\,h (1\,wk) $\to$ 12\,h \\
\bottomrule
\end{tabular}
\end{table*}

To evaluate all models, we report weighted mean absolute percentage error,
\[
\mathrm{WMAPE} = \frac{\sum_{t,i}|\hat{y}_{t,i} - y_{t,i}|}{\sum_{t,i}|y_{t,i}|},
\]
which allows comparison across datasets with different native units. It measures total absolute error as a share of the total observed value, so lower values are better. A graph-free GRU that forecasts each node from its own history serves as the primary baseline. All neural models use hidden dimension~64, learning rate $10^{-3}$, batch size~32, AdamW optimizer, SmoothL1 loss, and early stopping on validation WMAPE. Each configuration runs with seeds $\{7, 11, 23\}$; we report mean\,$\pm$\,std.

\subsection{Results and Analysis}
\label{sec:results}

\subsubsection{\rev{Does Graph Exposure Matter?}}

\rev{With the domain graph fixed, we compare the exposure strategies defined in Section~\ref{sec:exposure}. Table~\ref{tab:coupling} reports their forecast errors alongside the GRU baseline.}

\begin{table}[t]
\centering
\caption{Coupling results (domain graph fixed). WMAPE~(\%) mean\,\pmstd{std} over 3~seeds. \textbf{Bold}~= best learned model per dataset.}
\label{tab:coupling}
\small
\begin{tabular}{@{}lrrr@{}}
\toprule
\textbf{Coupling} & \textbf{Unemp.} & \textbf{Traffic} & \textbf{Elec.} \\
\midrule
GRU (no graph)          & 1.63\pmstd{.40} & 7.74\pmstd{.22} & 6.18\pmstd{.08} \\
GCN$\to$GRU             & 1.68\pmstd{.90} & 7.95\pmstd{.13} & 9.90\pmstd{2.74} \\
T-GCN                   & 1.42\pmstd{.73} & 8.09\pmstd{.09} & 10.43\pmstd{2.97} \\
GRUCell$\to$GCN         & 1.85\pmstd{.73} & 8.08\pmstd{.08} & 10.73\pmstd{2.99} \\
Late GRU$\to$GCN        & 1.57\pmstd{.33} & 7.96\pmstd{.12} & 9.21\pmstd{2.46} \\
Late + gate             & \textbf{1.36}\pmstd{.17} & \textbf{7.64}\pmstd{.14} & \textbf{5.27}\pmstd{.07} \\
\bottomrule
\end{tabular}

\end{table}

As Table~\ref{tab:coupling} shows, early and recurrent designs are harmful on traffic and electricity and unstable on unemployment. The effect is especially dramatic on electricity, where these models exceed 9\% WMAPE while the graph-free GRU achieves 6.18\%. The same graph, used differently, nearly doubles the error.

\rev{The comparison between the two late designs shows that delaying graph integration alone is insufficient. On electricity, late integration without a fusion gate has 9.21\% WMAPE, compared with 5.27\% with the gate and 6.18\% for the GRU baseline. On traffic, the corresponding errors are 7.96\%, 7.64\%, and 7.74\%. These results support the combination of late integration and selective fusion in these experiments. They do not establish that late integration is preferable independently of the gate.}

\rev{The same domain graph yields different forecast errors across integration designs. We next hold late gated fusion fixed to examine graph construction.}

\subsubsection{Does Graph Construction Matter?}

\rev{Table~\ref{tab:graph} compares the five construction rules from Section~\ref{sec:construction} using late gated fusion. The reference models are listed separately because their graph mechanisms and architectures also differ.}

\begin{table}[t]
\centering
\caption{Graph construction (late gated fusion) and reference methods. WMAPE~(\%) mean\,\pmstd{std} over 3~seeds. \textbf{Bold}~= best per dataset.}
\label{tab:graph}
\small
\begin{tabular}{@{}lrrr@{}}
\toprule
\textbf{Graph / Method} & \textbf{Unemp.} & \textbf{Traffic} & \textbf{Elec.} \\
GRU (no graph)          & 1.63\pmstd{.40} & 7.74\pmstd{.22} & 6.18\pmstd{.08} \\
\midrule
\multicolumn{4}{@{}l}{\emph{Late gated fusion:}} \\
\quad Domain prior      & 1.36\pmstd{.17} & 7.64\pmstd{.14} & 5.27\pmstd{.07} \\
\quad Random (deg.-matched) & 1.27\pmstd{.03} & 7.81\pmstd{.05} & 5.31\pmstd{.06} \\
\quad Raw-correlation kNN    & 1.25\pmstd{.07} & 7.48\pmstd{.17} & 5.11\pmstd{.03} \\
\quad Resid.-correlation kNN & \textbf{1.23}\pmstd{.09} & \textbf{6.87}\pmstd{.16} & \textbf{5.00}\pmstd{.04} \\
\quad Rewired 100\%          & 1.26\pmstd{.02} & 7.56\pmstd{.09} & 5.36\pmstd{.09} \\
\midrule
\multicolumn{4}{@{}l}{\emph{Reference methods (own graph machinery):}} \\
\quad GAT + gate, domain   & 1.38\pmstd{.32} & 7.92\pmstd{.04} & 5.05\pmstd{.21} \\
\quad AGCRN, learned       & 2.14\pmstd{.99} & 7.83\pmstd{.07} & 6.06\pmstd{.19} \\
\quad CaST, domain         & 1.74\pmstd{.40} & 7.86\pmstd{.13} & 6.85\pmstd{.38} \\
\bottomrule
\end{tabular}
\end{table}

Table~\ref{tab:graph} reveals that the ``obvious'' domain graph is not always the best graph for forecasting. The clearest example is \textbf{electricity}: the physical grid tells the model which wires connect, but electricity demand actually co-moves along weather zones and usage patterns, not grid wiring. A residual-correlation graph---built after removing common cross-node movement---captures this and achieves 5.00\% WMAPE, outperforming both the domain graph (5.27\%) and the graph-free GRU (6.18\%). In other words, the right graph is the one that encodes \emph{residual predictive signal}, not structural proximity.

The other two datasets complete the picture. On \textbf{unemployment}, all graphs---domain, random, data-driven---cluster within 1.23--1.27\%, and even random edges perform comparably. When a single national trend explains 93\% of the variation, topology-specific signal is weak. All graph variants, including random edges, converge to similar results. On \textbf{traffic}, the residual-correlation graph (6.87\%) clearly outperforms the domain graph (7.64\%) and the graph-free GRU (7.74\%), showing that even in a domain with physical propagation, a data-driven graph can better capture the predictive structure.

To test whether this pattern is resolved by more specialized graph machinery, we also compare with three recent methods that bring their own sophisticated graph mechanisms, including CaST \cite{xia2023deciphering} for causal disentanglement, GAT \cite{velickovic2018gat} for learned attention weights per edge, and AGCRN \cite{bai2020agcrn} for fully learned adjacency. None outperform our simple recipe---late gated fusion with a residual-correlation graph---on any dataset, reinforcing that getting the graph signal and coupling right matters more than architectural complexity.

\FloatBarrier
\section{Conclusion and Future Extensions}

\begin{revision}
We examined graph construction and graph exposure in three forecasting tasks. With each domain graph fixed, forecast error varied across integration designs. Late gated fusion had the lowest mean error among the tested designs, while late integration without a gate performed poorly on electricity. With late gated fusion fixed, the residual correlation graph had the lowest mean error among the tested graphs. Randomized graphs performed similarly on unemployment, so its results give limited evidence for the value of a specific topology. Together, these conditional comparisons show that both neighbor selection and graph integration matter when evaluating graph based forecasting models.
\end{revision}

This study also points to several extensions. We focus on three domains that cover common-shock, physical-propagation, and conditional graph-signal regimes, but other settings may involve different relational mechanisms, such as supply chains, mobility flows, weather systems, or social interaction networks. Our graph constructions are intentionally simple diagnostic probes, leaving room for richer adaptive, lagged, nonlinear, or exogenous-feature-aware graphs. Future work can apply the same construction--exposure decomposition to larger benchmark suites and to settings where the relevant graph changes over time.


\bibliographystyle{ACM-Reference-Format}
\bibliography{main}

\end{document}